%% file: ms.tex
\documentclass[10pt,twocolumn,letterpaper]{article}

\usepackage{cvpr}
\usepackage{times}
\usepackage{epsfig}
\usepackage{graphicx}
\usepackage{amsmath}
\usepackage{amssymb}
\usepackage{multirow}
\usepackage{cite}
\usepackage{algorithm, algorithmicx}
\usepackage{bm}
\usepackage{bbm}
\usepackage{booktabs}
\usepackage{url}
\usepackage{relsize}
\usepackage[noend]{algpseudocode}
\usepackage[pagebackref=true,breaklinks=true,letterpaper=true,colorlinks,bookmarks=false]{hyperref}

\usepackage{stfloats}
\usepackage{enumitem}
\usepackage{leftidx}
\setlist{nosep}
\cvprfinalcopy

\def\cvprPaperID{4001} 

\ifcvprfinal\fi

\newcommand{\x}{\mathbf{x}}

\newcommand{\y}{\mathbf{y}}

\begin{document}
\title{Fast Monte-Carlo Localization on Aerial Vehicles using \\Approximate Continuous Belief Representations}
\author{Aditya Dhawale\thanks{Contributed equally to this work.~Authors are with The Robotics Institute, Carnegie Mellon University, Pittsburgh, PA, 15213, USA. {\texttt{\{adityand, kshaurya, nmichael\}@cmu.edu}}}
	\and
	Kumar Shaurya Shankar\footnotemark[1]
	\and
	Nathan Michael
}
\maketitle
\thispagestyle{empty}
\begin{abstract}
	Size, weight, and power constrained platforms impose constraints on computational resources that introduce unique challenges in implementing localization algorithms. We present a framework to perform fast localization on such platforms enabled by the compressive capabilities of Gaussian Mixture Model representations of point cloud data. Given raw structural data from a depth sensor and pitch and roll estimates from an on-board attitude reference system, a multi-hypothesis particle filter localizes the vehicle by exploiting the likelihood of the data originating from the mixture model. We demonstrate analysis of this likelihood in the vicinity of the ground truth pose and detail its utilization in a particle filter-based vehicle localization strategy, and later present results of real-time implementations on a desktop system and an off-the-shelf embedded platform that outperform localization results from running a state-of-the-art algorithm on the same environment.
\end{abstract}
\section{Introduction}
\input{introduction}
\section{Approach}
\input{approach}
\section{Fast Localization}
\input{implementation}
\section{Results}
\input{results}
\section{Summary and Future Work}
\input{discussion}

{\small
	\bibliographystyle{ieee}
	\bibliography{references}
}
\end{document}

%% file: introduction.tex
\begin{figure*}[t!]
	\centering
	\includegraphics[width=0.8\linewidth]{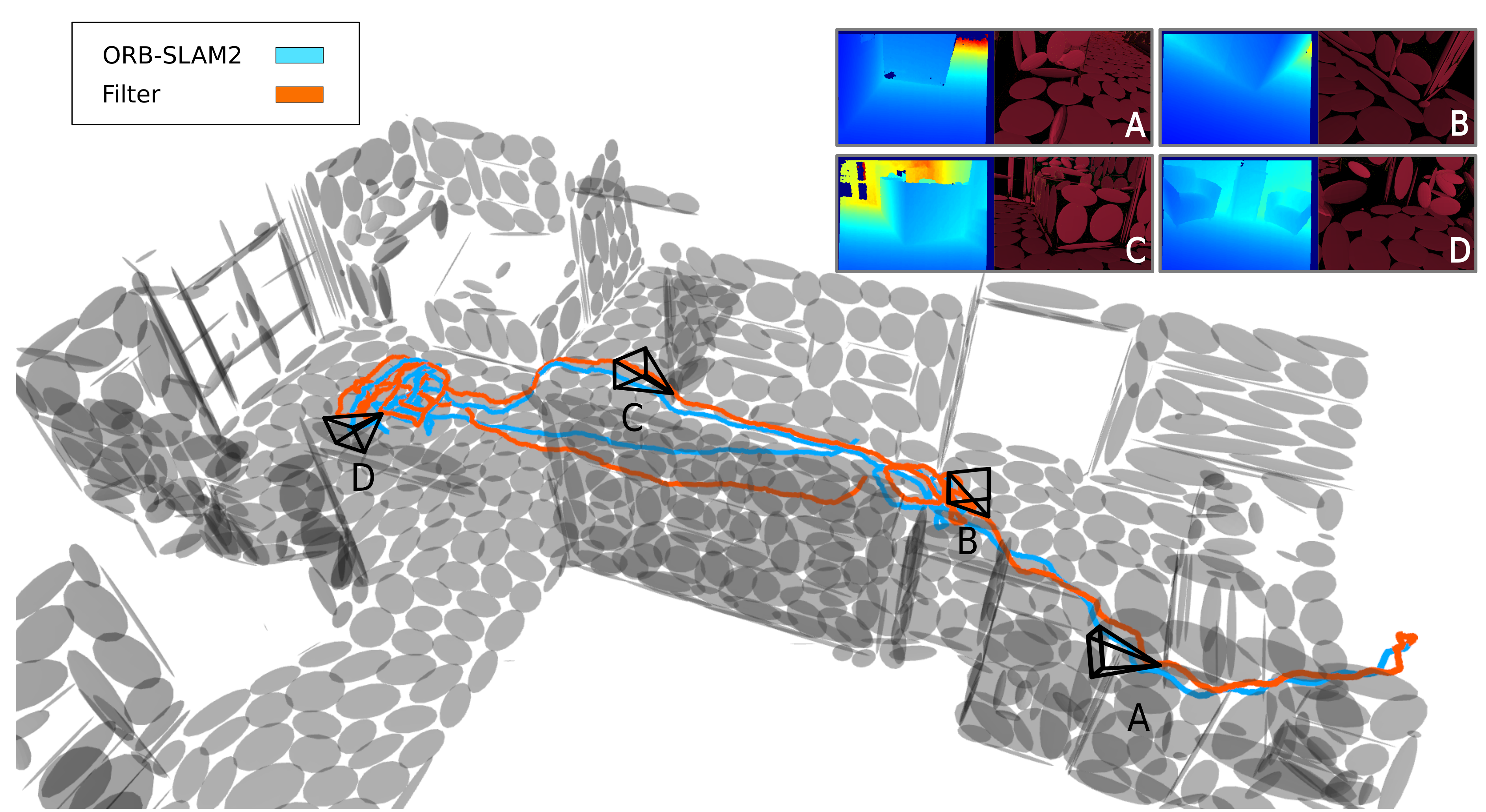}
	\caption{
		Comparison of the mean particle filter pose (orange) with that of the integrated process model trajectory (cyan) from a representative (office) dataset. The filter estimate is initialized with a uniform distribution away from the true location of the vehicle. As the camera observes more informative data the filter quickly converges to the correct pose. \textit{Top Right}: Four views of the raw point cloud sensor data and the corresponding view of the GMM map from the mean of the particle filter estimates. The GMM components are superimposed on top of the source point cloud with their $3\mathbf{\Sigma}$ bounds visualized as gray ellipsoids.}\label{fig:mockup}
\end{figure*}
For an agent lost in a known environment, much of the cost of localization can be offset by precomputing measures of what the sensor is expected to see; localization can then be cast as the much simpler problem of a search through these pre-existing ``hallucinated'' views. However, exhaustively considering all possible views incurs a prohibitive cost that increases exponentially with both the dimensionality of the state space and the size of the environment. Further, na\"{\i}ve pre-rendering approaches can be susceptible to errors caused by perceptual aliasing due to slight variations in the environment appearance or by regions that are not feature rich, such as blank walls~\cite{angeli2008fast}.

In this paper we present a framework enabling rapid computation of sensor data likelihood via an environment representation that is both high-fidelity and memory efficient. The framework models a depth camera observation as being sampled from the environment representation with a likelihood measure that varies smoothly about the true camera pose. We thus exploit the dramatically reduced storage complexity of the representation and the local spatial regularity of a fast-to-compute likelihood function to re-cast the pose estimation problem as that of Monte-Carlo localization~\cite{thrun2005probabilistic}.

Similar in spirit to our choice of representation,~\cite{fallon2012efficient} map the world with a succinct, albeit restrictive parameterization of using only dominant planes. Our choice of representation however, does not have any such restrictions. Likewise,~\cite{biswas2012depth} is similar in its restriction of representation to a structured world consisting of planes and edges and only localizes in 2D. Approaches such as NDT occupancy maps~\cite{stoyanov2012fast, oishi2013nd, das20133d}, unlike the representation used in this work, learn independent distributions over each discretized cell leading to a lower fidelity representation at the cell boundaries~\cite{srivastava2016approximate,eckart2015mlmd}. Further, for fast pointcloud alignment they require pre-computation of the incoming data for comparison with a stored model, known as Distribution-to-Distribution (D2D) registration. The more accurate Point-to-Distribution (P2D) registration approaches, however, are not as real-time viable~\cite{das20133d, magnusson2015beyond}, and are less so for our purpose. In~\cite{burguera2008likelihood} the authors represent each point in the reference scan of a 2D sonar scan as an isotropic Gaussian distribution and iteratively compute the 2D transformation that maximizes the likelihood of all points in the target scan. Our approach, on the other hand, represents clusters of points in the reference scan as individual anisotropic Gaussian components and is not restricted to transformations of small magnitude.

Closest in terms of our choice of implementation framework,~\cite{fang2015real} propose a particle filter based real-time RGB-D pose estimation approach and~\cite{bry2012state} present a real-time localization approach on a fixed-winged aircraft during aggressive flights with a laser scanner and an IMU. The latter work implicitly exploits much more restricted dynamics of a fixed wing aircraft within its process model and further, both these approaches use the OctoMap~\cite{hornung2013octomap} representation to provide correction updates within their filter estimates. Note that the memory footprint of an OctoMap is much greater than that of our choice of representation. For instance, memory consumption for an OctoMap of dataset D3 (Sec.~\ref{sec:experimental_setup}) with $0.1~\text{m}$ resolution is $872$ KB while the corresponding data usage for our Gaussian Mixture Model (GMM) map representation consisting of $1000$ Gaussian components is $40$ KB.


Prior works exploiting known map appearance for precise monocular pose estimation~\cite{stewart2012laps,pascoe2015robust,oksimultaneous,gonccalves2011real} employ a textured depth map within an iterative optimization framework to compute the warp that minimizes a photometric cost function between a rendered image and the live image such as the Normalized Information Distance\cite{pascoe2015robust}, that is robust to illumination change, or a Sum of Squared Differences cost function with an affine illumination model to tackle illumination change\cite{oksimultaneous}. Both algorithms rely on initialization for tracking via a GPS prior or an ORB-based bag-of-words approach, respectively, and expensive ray casted dense textured data for refinement.
%
Note that in contrast to the above mentioned algorithms that use RGB information, we only use depth observations and only project a finite number of mixture components as opposed to dense pre-rendered views of the map.

Our framework solves the problem of 6 Degree-of-Freedom pose estimation for Size, Weight, and Power (SWaP) constrained micro air vehicles operating in a known dense 3D pointcloud environment with an onboard monocular depth camera and Inertial Measurement Unit (IMU). We assume that the vehicle pitch and roll are obtained from an attitude estimation algorithm using the IMU in order to constrain the search space to just heading and position. Our main contributions are:
\begin{itemize}
	\item A particle filter-based localization strategy based on a high fidelity, memory efficient environment representation enabled by a fast likelihood computation approximation; and
	\item Experimental evaluation of the approach on a desktop and an off-the-shelf mobile GPU system.
\end{itemize}


%% file: approach.tex
Contemporary direct tracking algorithms require the projection of a large number of dense or semi-dense points into image space to align the current sensor data to a reference model. In contrast, we employ GMMs as a succinct parameterized representation to achieve orders of magnitude computational savings via an analytic projection into image space. The consequent reduction in complexity enables projection in multiple pose hypotheses concurrently in real-time and motivates this work.

This section details the choice of environment representation and how it enables the proposed real-time sensor data likelihood computation.

%
\subsection{Spatial GMMs as an Environment Representation for Tracking}
Conventional means of representing maps such as voxel grids discretize space to encode occupancy leading to resolution dependent model fidelity and memory efficiency. An alternate approach is to represent occupancy using a GMM map that attempts to approximate the underlying distribution from which sensor measurements are sampled. This formulation is capable of representing the environment model with as high a fidelity as required that scales gracefully with model complexity when used in a hierarchical fashion~\cite{srivastava2016approximate}. Additionally, this representation provides a probabilistic uncertainty estimate of the occupancy at any sampled location.
Fitting these models to data in real time is possible due to recent advances that enable efficient operation~\cite{eckart2016accelerated}.
We utilize the contribution presented in~\cite{srivastava2016approximate, eckart2016accelerated} to inform the number of Gaussian components required to pre-compute GMM maps of the environment point cloud at various fidelity levels. For the purpose of this paper, however, we limit the discussion to using only GMM maps at a certain fidelity level chosen according to Sec.~\ref{sec:sensitivity_analysis}, but note that the approach can be readily extended to a hierarchical formulation.
%
\begin{figure}[h]
	\centering
	\includegraphics[width=0.7\linewidth]{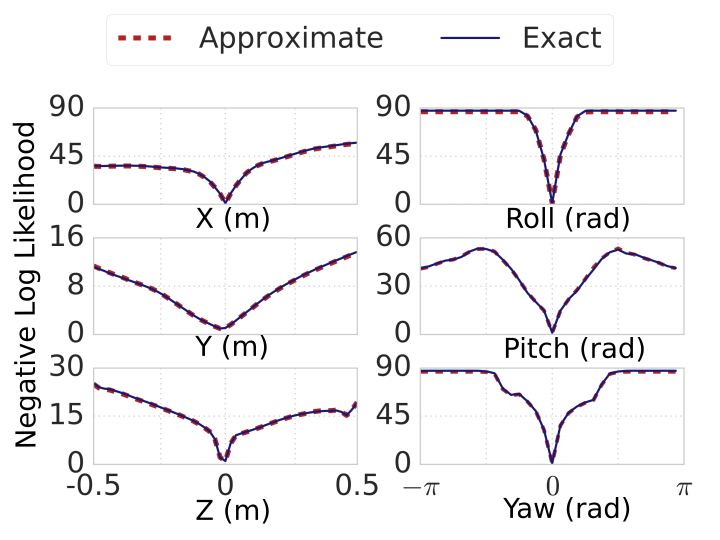}
	\caption{Negative log-likelihood plots of sensor data acquired from camera poses offset from a randomly chosen true pose in dataset D1 by incremental linear and rotational displacements. Utilizing only the relevant components using the approximation discussed in Sec.~\ref{sec:membership} leads to almost identical likelihoods as when utilizing all the Gaussian components present in the model.}\label{fig:likelihood}
\end{figure}

A spatial GMM represents the probability of matter existing at a specific position $\mathbf{P}^w$ given the model component parameters $\bm{\Theta} = \{\bm{\mu}_i, \mathbf{\Sigma}_i, \lambda_i\}_{i=1\ldots M}$ such that
\begin{equation}\label{eq:gmm}
	p\big(\mathbf{P}^w;\bm{\Theta}\big) = \sum_{i}^{M} \lambda_i \mathcal{N}( {\bm{\mu}_i}, \mathbf{\Sigma}_i)
\end{equation}
where $ \lambda_i $ is the mixture weight, ${\bm{\mu}_i}$ the mean 3D position, and $\bm{\Sigma}_i$ the covariance of the $i^{\text{th}}$ component of the GMM respectively, with $ \sum_{i}^M \lambda_i = 1 $ and $ \lambda_i > 0 $.%
\subsection{Projection of a GMM Component into Image Space}\label{sec:projection}
In order to determine relevant mixture components of a given spatial GMM map for evaluating sensor data likelihood we first analytically project the mixture components into image space.

For a point $\mathbf{P}^w$ in the world frame, the transformed position $\mathbf{P}^c $ in a camera frame $ \mathbf{T}_w^c $ is denoted as
\[ \mathbf{P}^c = \mathbf{R}_w^c \mathbf{P}^w +  \mathbf{t}_w^c\]
where $  \mathbf{R}_w^c $ and $ \mathbf{t}_w^c $ are the corresponding rotation matrix and translation vectors, respectively. Since this is a linear operation on $ \mathbf{P}^w$, using Eq.~\ref{eq:gmm} the transformed distribution of points in the camera frame for the $ i^{\text{th}} $ component is
\[p\big(\mathbf{P}^c;\bm{\Theta}_i\big) = \mathcal{N}(\mathbf{T}_w^c\bm{\mu}_i,\mathbf{R}_w^c \mathbf{\Sigma}_i {\mathbf{R}_w^c}^\text{T})\]
Consider a sample $ \x_s \sim \mathcal{N}(\bm{\mu}, \bm{\Sigma}) $ and a monotonic continuous nonlinear function $ \y = f(\x) $ (where $ \y $ and $ \x_s $ are in the same space). The first order Taylor series expansion about a point ${\x_s}$ leads to
\[ \y \sim \mathcal{N}\left(E[f({\x})], \left. \frac{\partial f}{\partial \x}\right|_{\x = {\x_s}} \mathbf{\Sigma} \left. \frac{\partial f}{\partial \x}\right|_{\x = {\x_s}} ^\text{T} \right) \]

Under the standard pinhole projection model, a point $\mathbf{P}$ in the camera frame is projected to the image space using the operation $ \pi : \mathcal{R}^3 \rightarrow \mathcal{R}^2 $ defined as
\[
	\bm{\pi}(\mathbf{P}) = \left[ \begin{array}{c} c_x + f \frac{\mathbf{P}_x}{\mathbf{P}_z} \\
			c_y + f \frac{\mathbf{P}_y}{\mathbf{P}_z}\end{array} \right] \]
where $ f $ is the focal length of the camera, and $c_x$ and $c_y$ are the principal point offsets. The derivative of the projection operation with respect to the 3D point $ \mathbf{P}$ is
\[
	\frac{\partial \bm{\pi}}{\partial \mathbf{P}} = \left[ \begin{array}{ccc} \frac{f}{\mathbf{P}_z} & 0                      & -f \frac{\mathbf{P}_x}{\mathbf{P}_z^2} \\
			0                            & \frac{f}{\mathbf{P}_z} & -f \frac{\mathbf{P}_y}{\mathbf{P}_z^2}\end{array} \right]
\]
Thus the projection of a 3D normal distribution component into image space is the 2D normal distribution
\begin{equation}\label{eq:proj}
	\begin{split}
		&p(u,v; \bm{\Theta}_i) =\\&\mathcal{N}
		\left( \bm{\pi}(\mathbf{T}_w^c{\bm{\mu}_i}),
		\left. \frac{\partial \bm{\pi}}
		{\partial \mathbf{P}}\right|_{\mathbf{T}_w^c{\bm{\mu}_i}}
		\mathbf{R}_w^c
		\mathbf{\Sigma}_i
		{\mathbf{R}_w^c}^\text{T}
		\left. \frac{\partial \bm{\pi}}
		{\partial \mathbf{P}}^\text{T} \right|_{\mathbf{T}_w^c{\bm{\mu}_i}} \right)
	\end{split}
\end{equation}
where $ u,v $ are pixel coordinates in the image.
\subsection{Estimating the Likelihood of a Camera Pose Hypothesis}\label{sec:approx_like}
As shown in the previous subsection, each Gaussian component can be projected into image space as a 2D Gaussian distribution. We utilize this property to determine relevant components for computing the likelihood of sensor data~(Sec.~\ref{sec:membership}). Given a scan $\mathbf{Z}_t$ of depth pixels $\{z_1, z_2,\dots,z_k\}$ from a sensor scan and a set of 3D GMM parameters~$\bm{\Theta}$, the log likelihood of the scan being sampled from the GMM is defined as
\begin{equation}
	\begin{split}
		&l\big(\mathbf{Z}_t|\mathbf{\Theta}, \mathbf{T}_w^c\big) =\\
		&\sum_i^K \ln \sum_j^M \mathbbm{1}_j\lambda_j \mathcal{N}\big(\bm{\pi}^{-1}(z_i);\mathbf{T}_w^c\bm{\mu}_j,\mathbf{R}_w^c \mathbf{\Sigma}_j {\mathbf{R}_w^c}^\text{T}\big)
		\label{eq:approx_likelihood_3d}
	\end{split}
\end{equation}
where $\mathbbm{1}_i$ is a binary indicator function that signifies if the $i^{\text{th}}$ component is used to compute the log likelihood, $ \bm{\pi}^{-1} $ is the inverse projection from depth image pixel to 3D points, and $K$ is the number of pixels in the sensor scan. This likelihood should peak at the true sensor pose and decay smoothly in the local neighbourhood, which is indeed observed as shown in Fig.~\ref{fig:likelihood}.

\begin{figure}[h]
	\centering
	\includegraphics[width=0.9\linewidth]{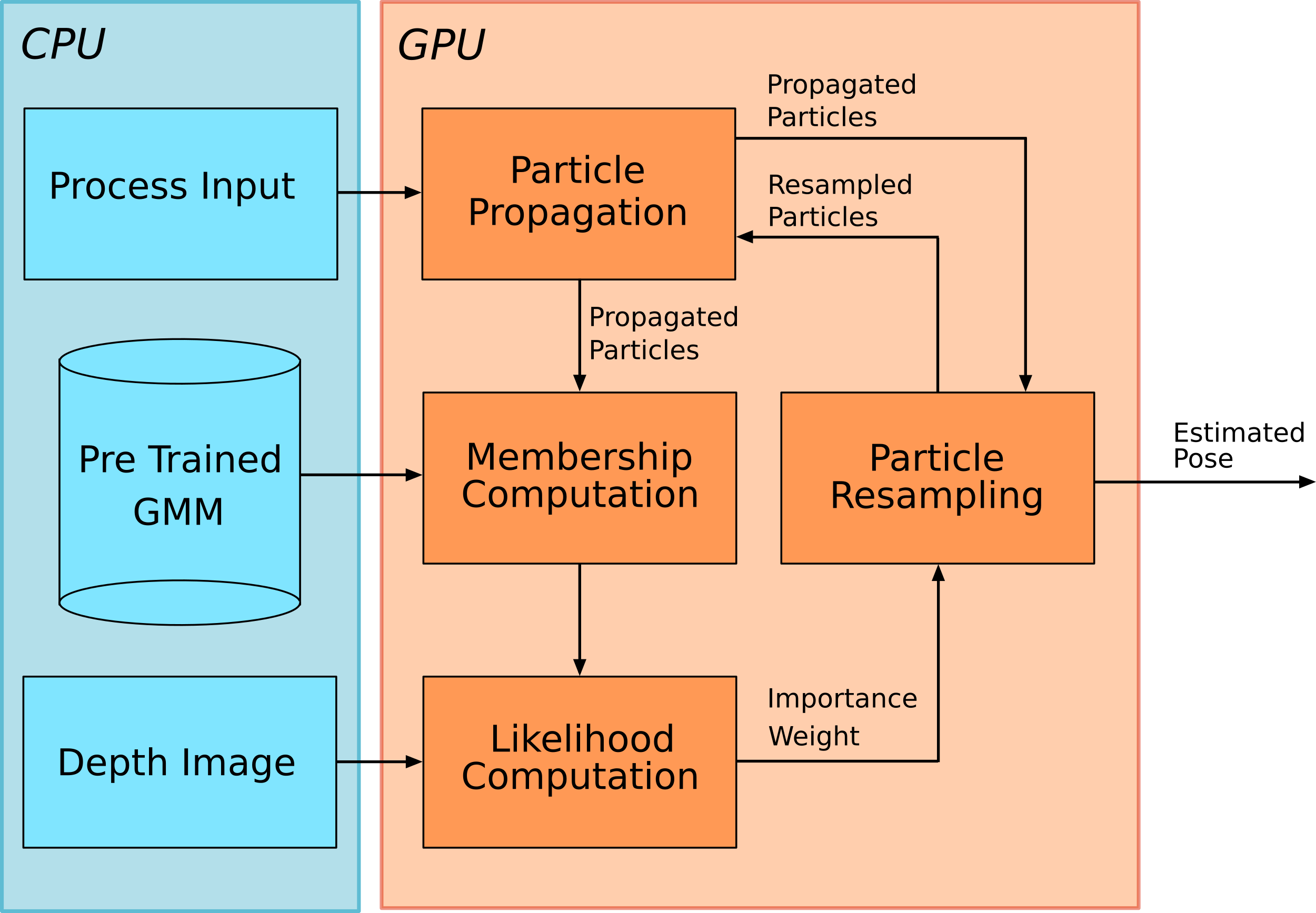}
	\caption{System overview. The algorithm operates on depth image streams and a source of odometry given a precomputed GMM map of the environment. For each particle, the GMM components are projected into image space using its current pose hypothesis. Relevant components are sub-selected and are then used to compute the likelihood of the depth image. The likelihood values for all the particles are used to resample a new set of particles that are then forward propagated using the process model.}
\end{figure}
\subsection{Tracking Multiple Hypotheses}
The discussion above only considers the nature of the likelihood in the vicinity of the true location; in practice it is not reasonable to assume that a single viewpoint suffices to localize the system as perceptual aliasing may arise due to a paucity of data that precludes state observability. Hence, we require a technique that permits tracking of multiple hypotheses and ensures appropriate weighting of equally likely viewpoints given the current sensor observations.

A standard approach to tracking multiple hypotheses is a Monte Carlo filter (or particle filter). Particle filters operate by continuously sampling candidate particle poses and measure the likelihood of the current sensor data having originated at the sampled pose. Based on the relative scores of the samples the particles are resampled and propagated based on a process model (often a noisy source of odometry). Convergence is generally achieved as soon as the sequence of observations made over time render alternate hypotheses inadmissible. Note that due to their inherent structure particle filters are extremely parallelizable and we exploit this in our implementation.
\subsubsection{State Propagation}
We assume the presence of some odometry to drive the first order Markov process model and inject Gaussian noise into it. Note that we assume that we know the pitch and roll that can be obtained from the attitude and heading reference system onboard a robotic system to a high level of accuracy.
\subsubsection{Importance Weight}
The importance weight of a particle in the filter represents a score of how well the sensor scan matches the GMM map at its location. Since the negative log likelihood of the current scan $\mathbf{Z}_t$ being drawn from the GMM map is a minimum at the true location, as shown in Fig.~\ref{fig:likelihood}, in practice we use the inverse of the negative log likelihood. Thus, given the current state estimate ${\mathbf{T}_w^{c^{(i)}}}$ of a particle $ i $ out of $ N $ particles at time step $ t $, the corresponding normalized importance weight is
\begin{equation}\label{eq:weight}
	w_t^{(i)} = \frac{l\big(\mathbf{Z}_t|\mathbf{\Theta}, {\mathbf{T}_w^{c^{(i)}}}\big)^{-1}}{\sum_{j}^N l\big(\mathbf{Z}_t | \mathbf{\Theta}, {\mathbf{T}_w^{c^{(j)}}}\big)^{-1}}
\end{equation}

\subsubsection{Sampling Strategy}
A particle filter should ideally converge to the correct hypothesis after running for a finite amount of iterations with a reduction in the filter variance signifying the confidence of the filter. At the same time, an early reduction in the filter variance may cause the filter to diverge to an incorrect hypothesis and never recover due to low variance. In order to avoid such situations, we implement the stratified sampling strategy~\cite{kitagawa1996monte} in combination with low variance sampling~\cite{thrun2005probabilistic}. The particles are divided into random groups of equal weights and in each group we employ low variance sampling. This approach has low particle variance~\cite{thrun2005probabilistic} and works well when the particle filter is tracking multiple hypotheses at once.
\subsubsection{Handling Particle Deprivation}
One of the most common failure modalities of a particle filter is that of particle deprivation~\cite{van2001unscented}. Even with a large number of particles, the stochasticity intrinsic to a particle filter might cause it to diverge from the correct state.
We employ a modified version of Augmented MCL strategy as described in~\cite{thrun2005probabilistic} where instead of adding new particles we reinitialize $N_{\text{modify}}$ number of particles randomly selected from the original set using the parameters $\alpha_\text{slow}$ and $\alpha_\text{fast}$. This is done since we cannot increase the number of particles once the filter is initialized because of implementation limitations. For our process model we use diagonal covariances for translation, and the final choice of parameters in all our experiments is shown in Table~\ref{table:parameters}.
%
%

%% file: implementation.tex
\label{sec:membership}In order to perform fast localization using the above approach it is essential to compute the likelihood of the data given a proposed pose as quickly as possible.
Eq.~\ref{eq:approx_likelihood_3d} suggests that computing the likelihood of a scan having been sampled from the GMM map is the summation of the contribution of all the components within the GMM. However, the key insight here is that not all the components have a significant contribution to the likelihood.

The point clouds that we use in our experiments have roughly uniform coverage of points across the scene. As a consequence, all Gaussian components fit to these pointclouds end up having roughly equivalent mixture weight probabilities. This fact, in addition to the diminishing probability mass of the Gaussian distribution, permits the approximation of using only the projected components within spatial proximity of a certain pixel location for computing the likelihood of the corresponding 3D point being sampled from the map.
As an added optimization step we perform this membership computation over subdivided patches of the image. These optimizations have negligible effect on the computed likelihood value of the sensor data, as demonstrated in Fig.~\ref{fig:likelihood}.

\begin{figure}[h]
	\centering
	\includegraphics[width=\linewidth]{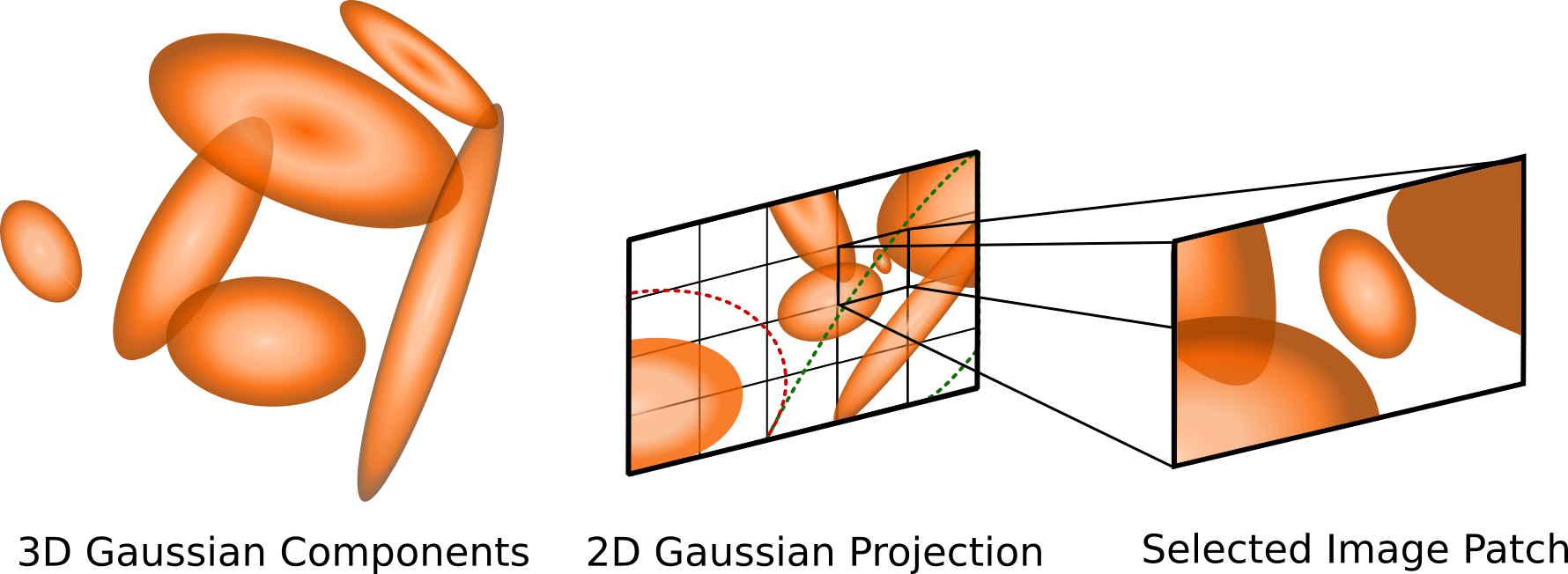}
	\caption{Membership computation process. 3D Gaussian components from the GMM representation of the world are projected to the image plane. The image is subdivided into multiple patches, where for a selected patch the relevant Gaussian members are determined for computing the likelihood. In order to determine the latter, we employ the heuristic described in Sec.~\ref{sec:membership}. For instance the inflated bounds of the bottom left projected component (red) do not contain the center of the selected patch; in contrast those of the bottom right (green) do, and the component is thus selected for computing the likelihood of data within that particular patch.}
	\label{fig:membership}
\end{figure}

We follow the following steps (graphically illustrated in Fig.~\ref{fig:membership}) to obtain the relevant components for computing the likelihood of a depth image:
\begin{itemize}
	\item Divide the image into $32\times32$ pixel patches;
	\item Compute the 2D projection of each Gaussian component on to the image plane of the depth sensor;
	\item Inflate the $3\mathbf{\Sigma}$-bound ellipse of the projected 2D Gaussian of each component by half the diagonal of the patch along its major and minor axis to generate ellipses $\mathcal{E}_i$; and
	\item For each patch, check if the center of the image patch $\mathit{c}_p$ lies within or on each of the $\mathcal{E}_i$ and update the indicator variable $\mathbbm{1}_{i, p}$ accordingly.
\end{itemize}
\begin{equation}
	\mathbbm{1}_{i, p} =
	\begin{cases}
		1 \text{, if } \mathit{c}_p \in \mathcal{E}_i \\
		0 \text{, otherwise}
	\end{cases}
\end{equation}
Given a set of updated indicator variables $\mathbbm{1}_{i,p}$ for all the Gaussian components $\mathbf{\Theta}$ and a depth image, $\mathbf{Z}_t$, the likelihood of the image can be computed as the sum of the likelihoods of all the image patches computed according to Eq.~\ref{eq:approx_likelihood_3d}.

%% file: results.tex
\subsection{Experiment Design}\label{sec:experimental_setup}
This section presents performance analysis of our filtering approach on a variety of datasets. First, we conduct a sensitivity analysis to determine the number of particles and the number of components we use in our implementation. Second, we analyze metric accuracy of the proposed filter on publicly available datasets and show that our filter output is consistent with ground truth. Third, we compare the localization performance of our approach with a state-of-the-art RGB-D tracking algorithm (ORB-SLAM2~\cite{mur2017orb}) on the same sequences and demonstrate superior performance for localization. Fourth, we demonstrate the ability of our approach to incorporate both different odometry algorithms and ground truth map acquisition methodologies. Finally, we analyze runtime performance of our filter and show that its runtime is competitive both on a desktop class system and on an embedded platform, thus enabling SWaP constrained operation.

We evaluate our approach on
\begin{itemize}
	\item D1: The~(a)~lounge and~(b)~copyroom datasets~\cite{zhou2013dense};
	\item D2: The voxblox dataset~\cite{oleynikova2016voxblox};
	\item D3: A representative dataset collected in-situ; and
	\item D4: The TUM Freiburg3 dataset~\cite{sturm12iros} for demonstrating the ability to generalize.
\end{itemize}
In all cases we utilize a fixed number of components (Sec.~\ref{sec:sensitivity_analysis}) to first fit a GMM to the pointcloud using the scikit-learn\footnote{\url{http://scikit-learn.org/stable/modules/mixture.html}} toolkit.

We employ two processing systems for evaluation: (1) A desktop with an Intel i7 CPU and an NVIDIA GTX 960 Ti GPU, and (2) An embedded NVIDIA TX2 platform.

\subsection{Sensitivity Analysis}
\label{sec:sensitivity_analysis}
Particle filters can achieve increased performance with large number of particles at the cost of increased computational complexity. Conversely too few particles can lead to divergence from the true location due to an inability to represent the true underlying distribution. In order to find the appropriate number of particles that ensure precision while still being computationally feasible we compare the filter performance with various number of particles against a ground truth filter with $ N = 16200$. Assuming the underlying distribution represented by the particle set to be a unimodal Gaussian (a valid assumption after convergence), we compute the variance of the KL-Divergence~\cite{hershey2007approximating} of multiple runs of the filter output with that of the ground truth filter to determine the empirically optimal parameters to be used in our implementation. A low value of the KL-Divergence variance indicates similar performance to the ground truth filter.

The fidelity of a GMM map to the underlying distribution monotonically increases with the number of components. However, the marginal benefit (in a KL-Divergence sense) of increasing the model complexity diminishes rapidly after adding an adequate number of components~\cite{srivastava2016approximate}. In order to determine the appropriate model complexity to represent the original map concisely while enabling accurate filter performance, we perform similar experiments with the optimal number of particles obtained from the previous study, this time with varying number of Gaussian components. \\
We compute the optimal parameters to be $N=1068$ and $M=1000$ based on D3, the dataset with the largest volumetric span. This specific parameter choice is further motivated by implementation constraints.
\begin{figure}[h]
	\centering
	\includegraphics[width=0.45\linewidth]{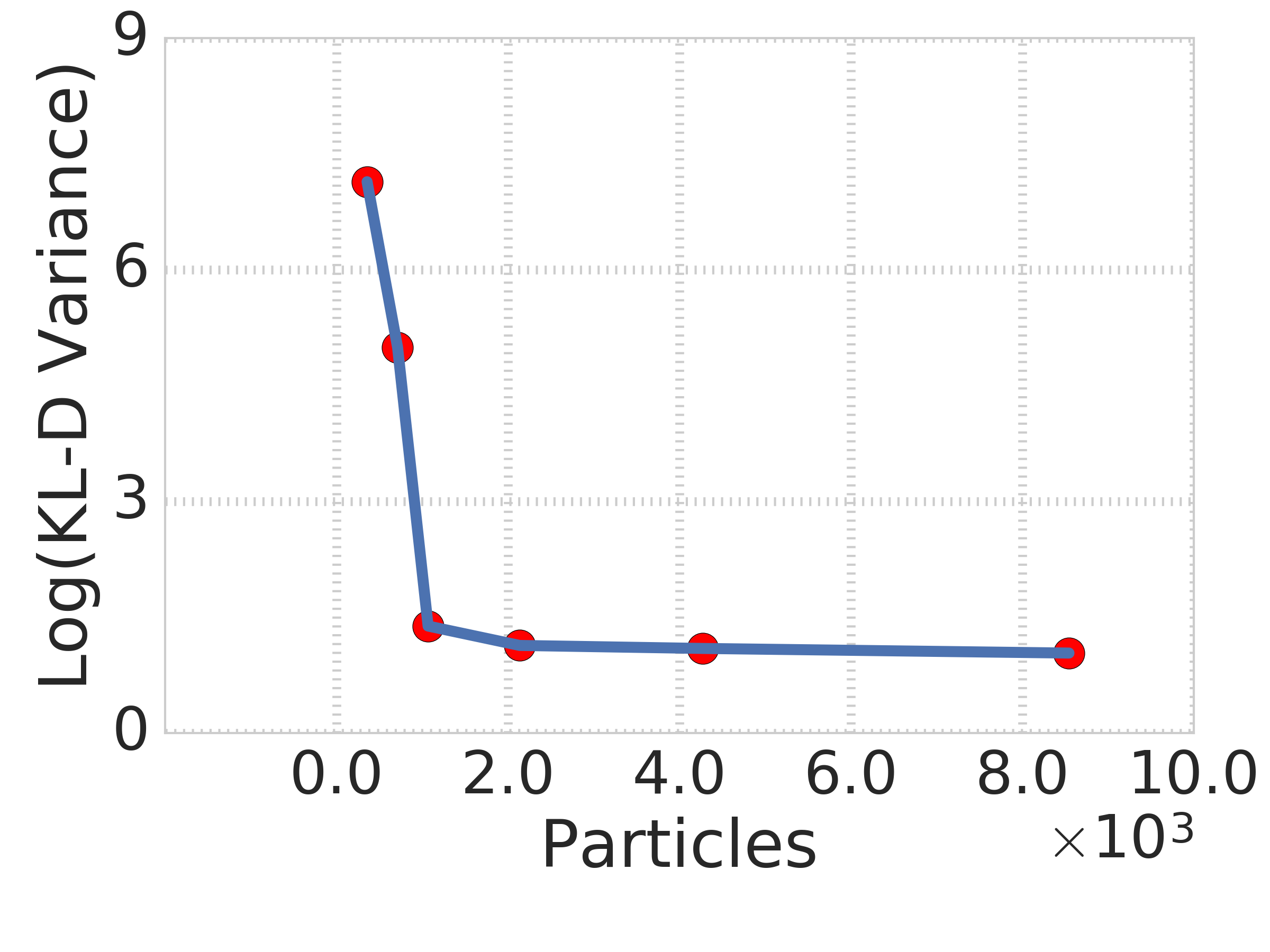}~
	\includegraphics[width=0.45\linewidth]{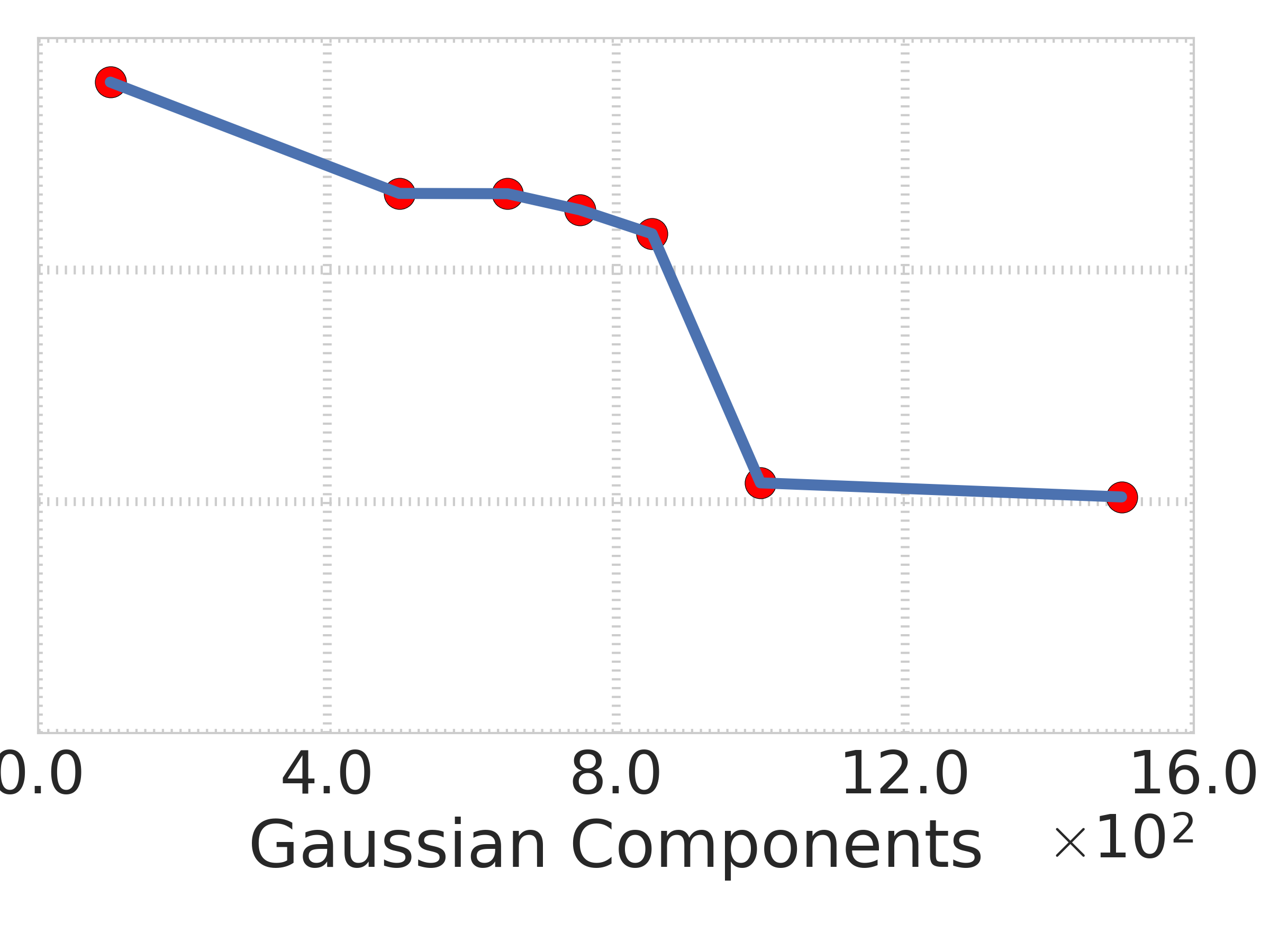}
	\caption{\textit{Left}: Log of variance of KL-Divergence between the ground truth filter $(N=16200)$ and filters with reduced particle counts. The knee point implies similar performance to the ground truth filter at particles counts $ N>1000 $. \textit{Right}: A similar comparison given a ground truth map with many components ($M=2500$) and those with a reduced number motivates the choice of $M>=1000$. Evaluated on D3.}\label{fig:convergence_plots}
\end{figure}
\begin{table}[t]
	\centering
		\caption{Filter hyperparameters}
	\scalebox{0.9}{
		\begin{tabular}{lcccc}
			\toprule
			 & \multicolumn{2}{c}{Process Noise $\sigma$} & \multirow{2}{*}{$ \alpha_{slow} $} & \multirow{2}{*}{$ \alpha_{fast} $}   \\
			 & Translation (m)                            & Yaw (rad)                          &                                    & \\
			\midrule
			{Desktop}
			 & 0.02
			 & 0.01
			 & 0.01
			 & 0.001                                                                                                                  \\
			{TX2}
			 & 0.025
			 & 0.1
			 & 0.05
			 & 0.005                                                                                                                  \\
			\bottomrule
		\end{tabular}}\label{table:parameters}
\end{table}

\subsection{Metric Accuracy Analysis}
In this subsection we discuss the localization accuracy of our approach. As mentioned in Sec.~\ref{sec:membership} since we do not add new particles when the filter observes particle deprivation and instead randomly reinitialize the particles from the original set, the Root Mean Squared Error (RMSE) of the filter estimate increases when the filter observes particle deprivation. This is highlighted in the plots as vertical shaded regions. For all our evaluations we run the filter 10 times on each dataset and report the average of the mean filter estimate. We do not quantify the sensitivity of the likelihood values to the AHRS pitch and roll estimates as they are accurate enough to not cause any significant difference.
\subsubsection{Evaluation with Ground Truth Datasets ({D1}, {D2})}
The objective of using these datasets is to demonstrate the ability of the filter to converge to the ground truth given perfect odometry. We generated a GMM map of the environments using the reconstructed point cloud and used the delta transforms between two consecutive reported sensor poses with added noise as our process model. In all these experiments, we initialized the particles from a uniform distribution over a $4~\text{m}$ cube and $\pi$ radians yaw orientation around the known initial location. D1(a) and D1(b) contain nominal motion of the sensor, while D2 consists of very aggressive motion in all degrees of freedom.

\begin{figure}[h]
	\centering
	\includegraphics[width=0.4\linewidth]{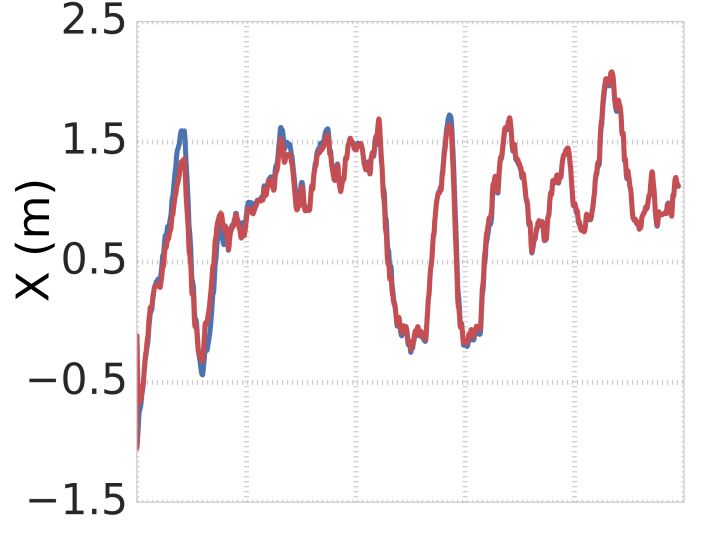}~\includegraphics[width=0.4\linewidth]{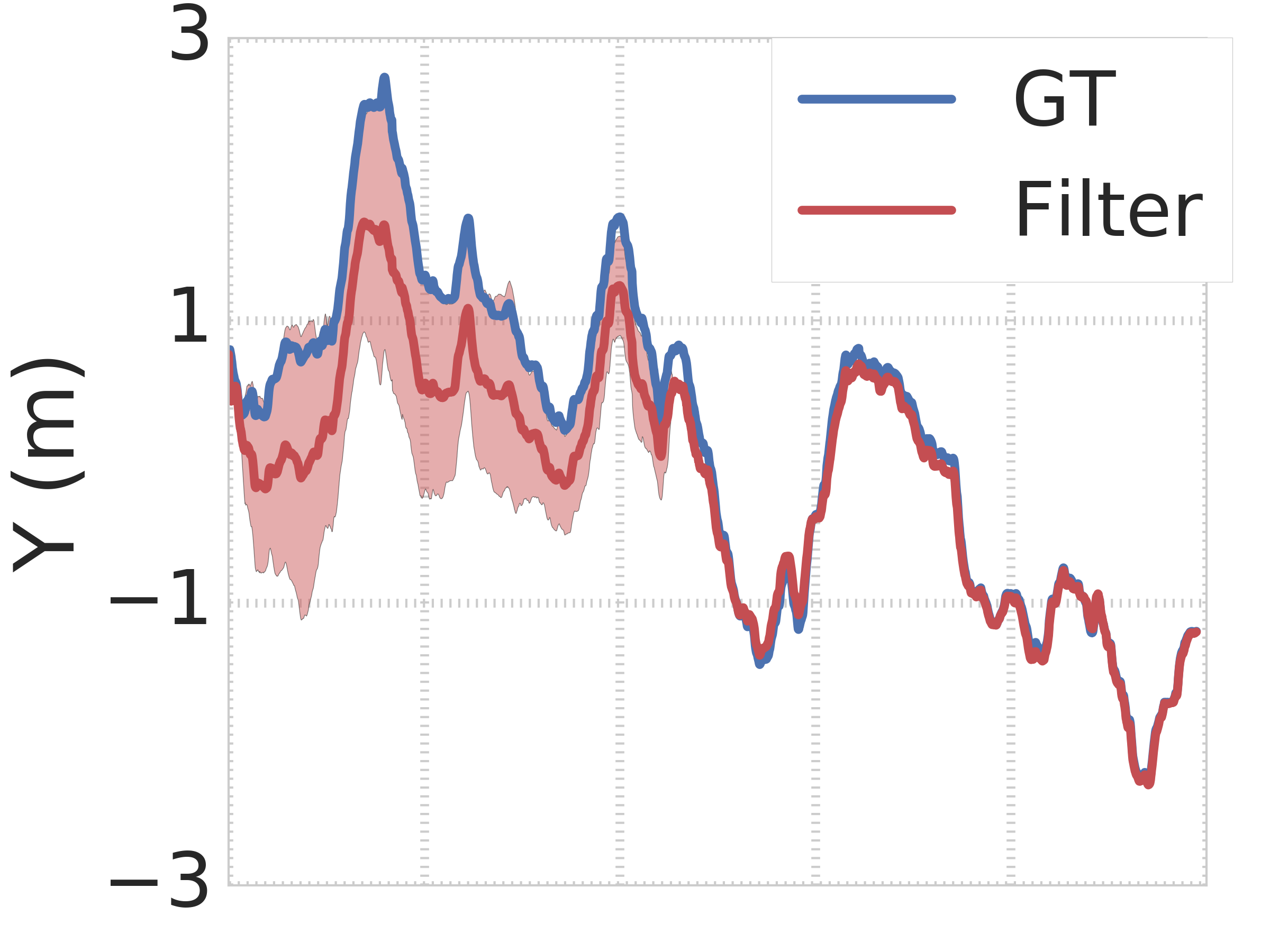}\\
	\includegraphics[width=0.4\linewidth]{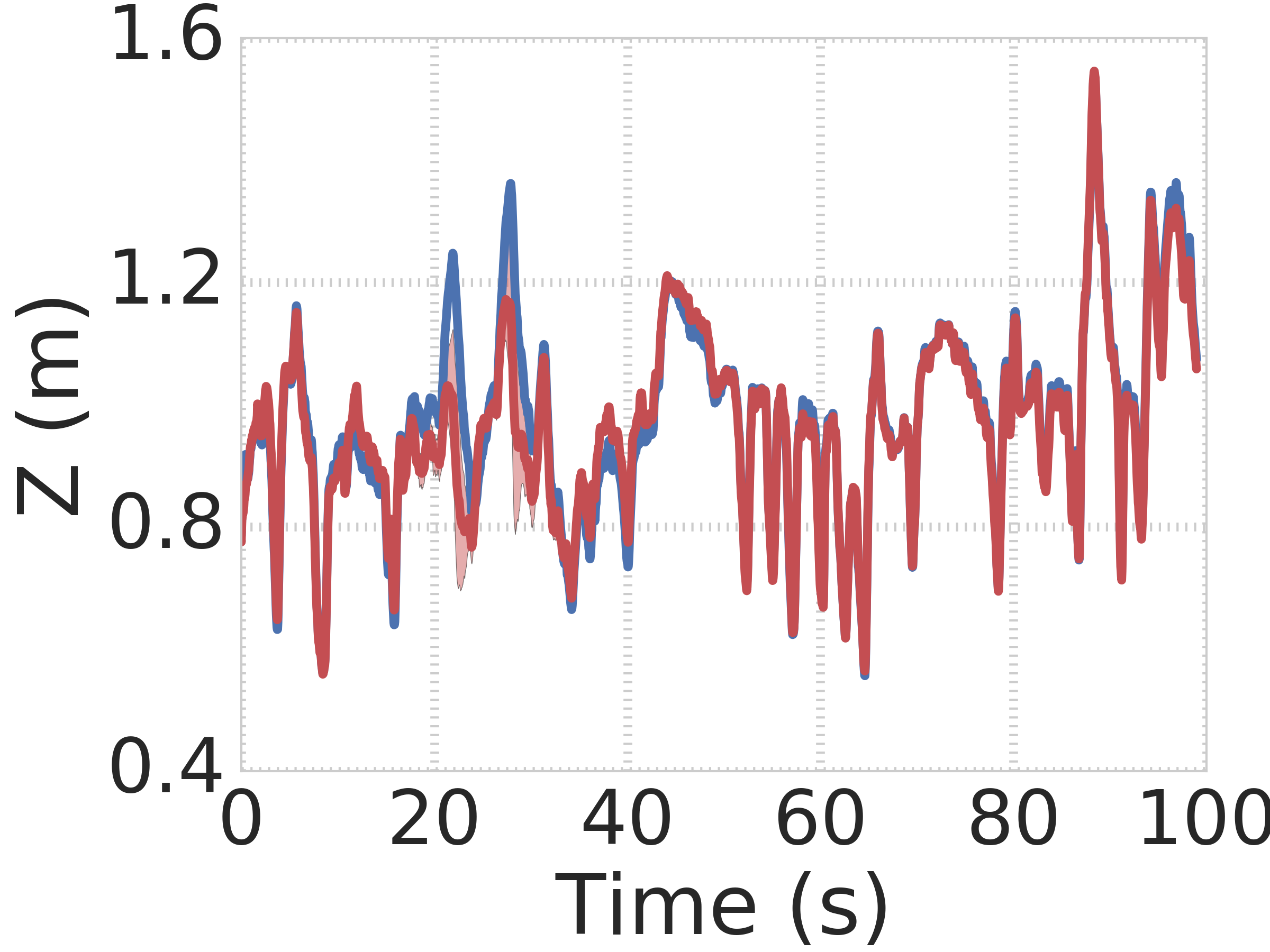}~\includegraphics[width=0.4\linewidth]{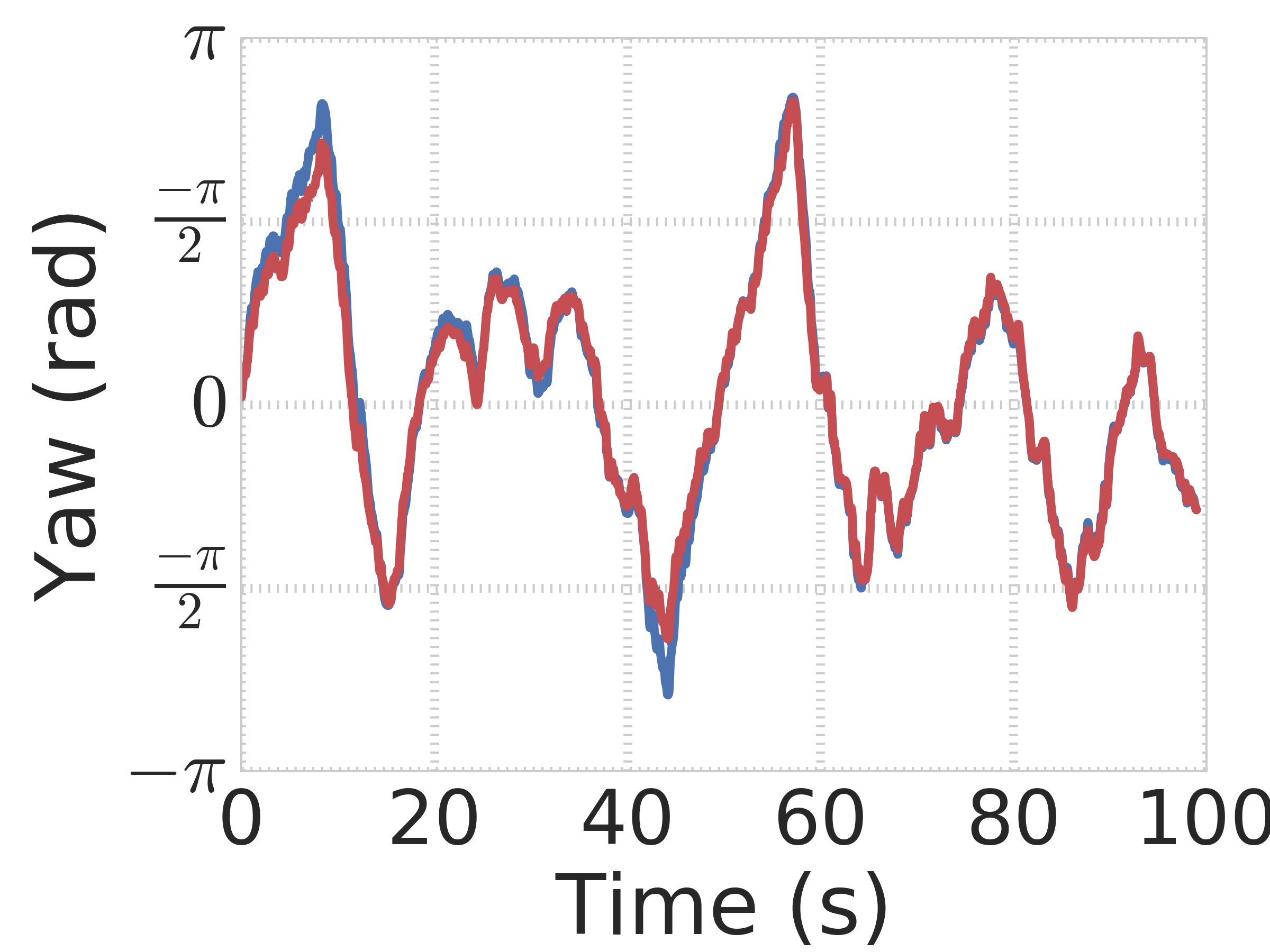}
	\caption{Mean trajectory (red) of the particle filter estimate of 10 trials on the D1(a) dataset compared to the process model trajectory (blue). The shaded region around the mean trajectory shows the variance of the filter estimate over multiple runs. The filter estimates have high variance in the beginning of the trajectories, but soon converge to the correct location and track the ground truth trajectory (blue).}\label{fig:stanford_lounge_traj}
\end{figure}

\begin{figure}[h]
	\centering
	\includegraphics[width=0.98\linewidth]{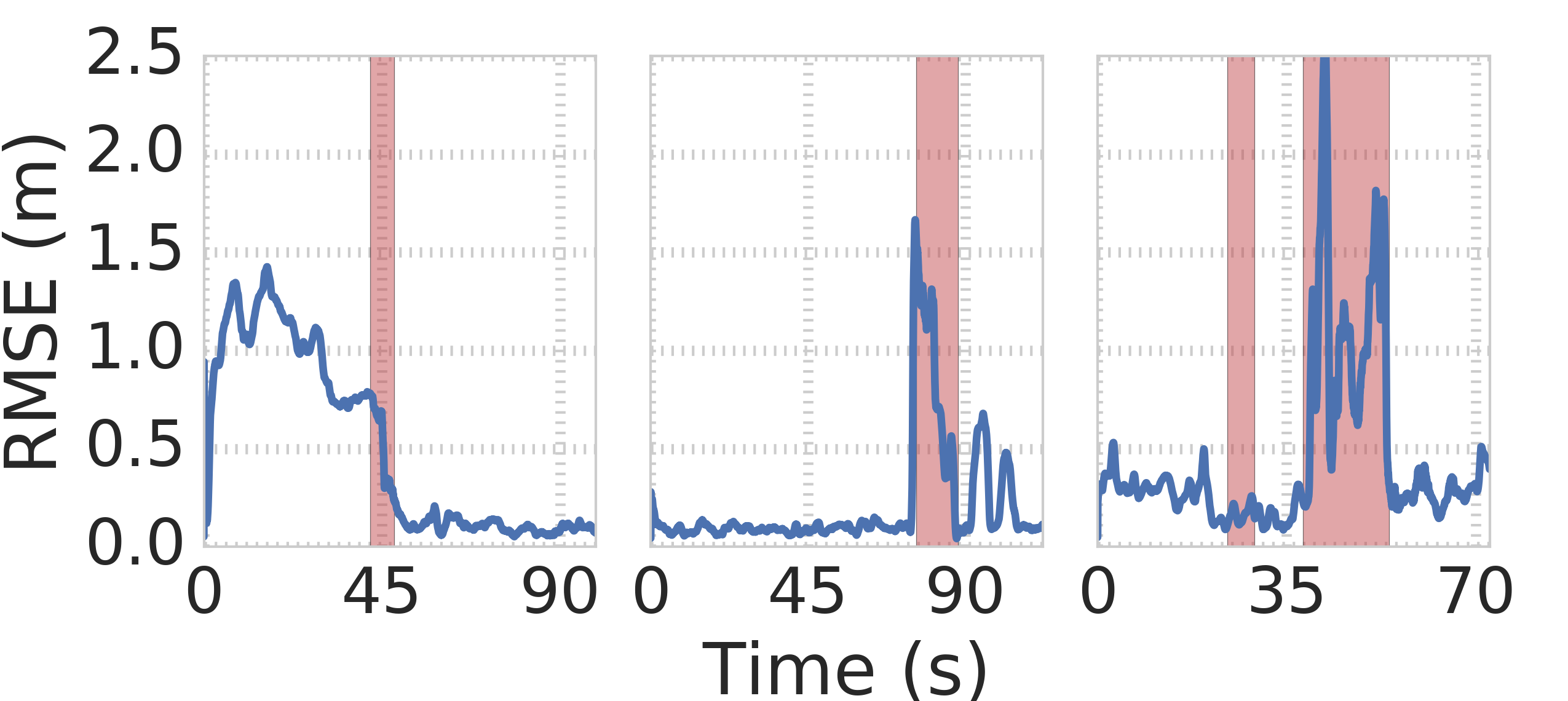}
	\caption{RMSE of 10 trials of the particle filter on  the D1(a), D1(b), and D2 datasets respectively. The region in red indicates the time at which the particle filter observes particle deprivation and a consequential RMSE rise.}\label{fig:RMSE_D1_and_2}
\end{figure}

The filter estimate converged to an incorrect hypothesis for some runs in the initial iterations due to the highly symmetric nature of the environments about the X axis, as can be seen in Fig.~\ref{fig:stanford_lounge_traj}. The RMSE of the filter poses for these datasets is presented in Fig.~\ref{fig:RMSE_D1_and_2}.
\vspace{-10pt}
\subsubsection{Evaluation with Representative Dataset (D3)}
The objective of using this dataset is to demonstrate results on a real-world application of the filter. We no longer use ground truth odometry. Additionally, since we don't have a baseline algorithm to directly compare against, we compare the localization performance against ORB-SLAM2 which builds its own succinct map representation. Note that ORB-SLAM2 also utilizes the RGB image data in the dataset whereas we only use the depth. Finally, we also briefly contrast the performance of the filter on the same dataset.

\begin{figure}[h]
	\centering
	\setlength\tabcolsep{0pt}
	\begin{tabular}{cc}
		\centering
		\includegraphics[width=0.45\linewidth]{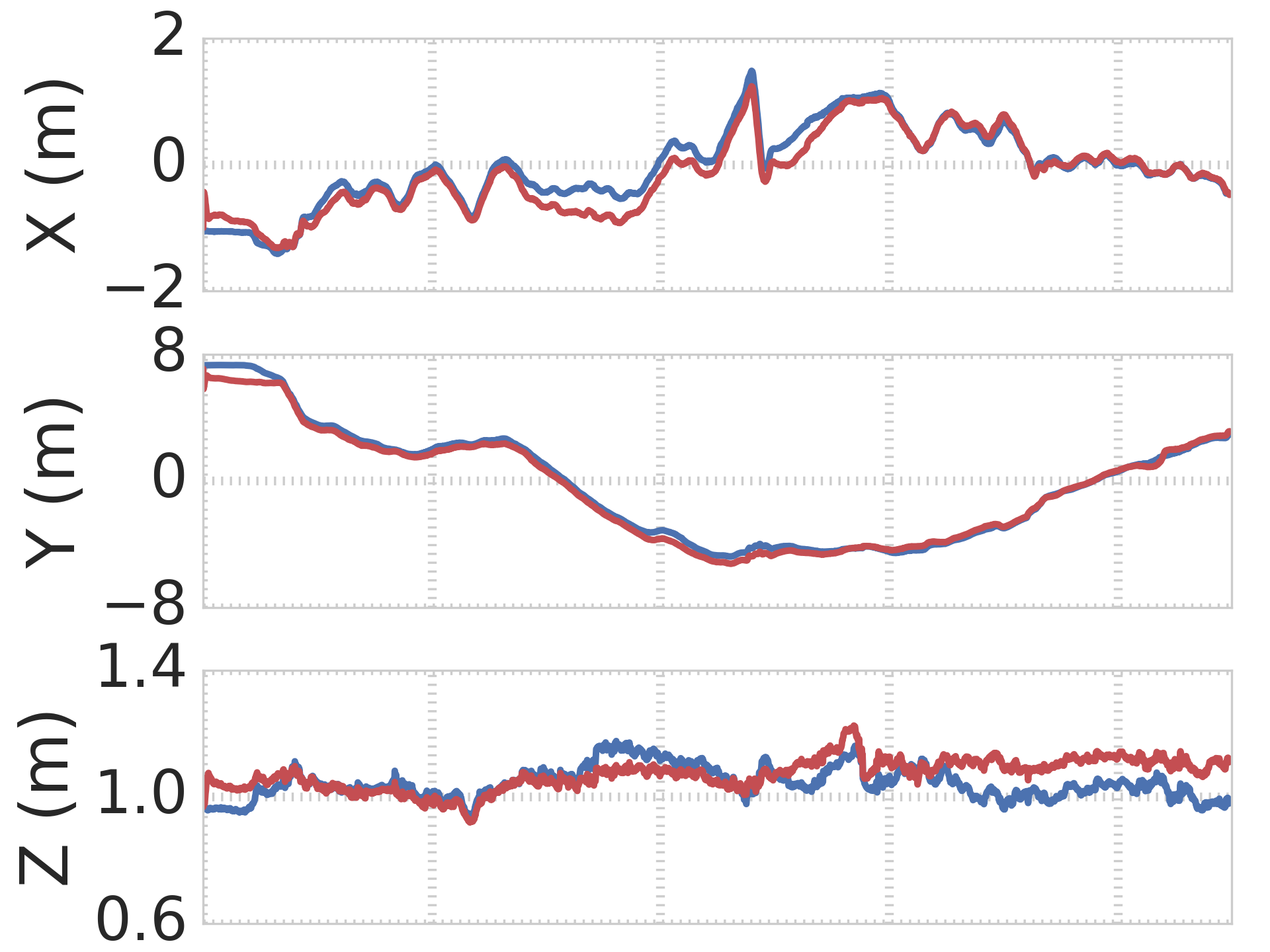} &
		\includegraphics[width=0.45\linewidth]{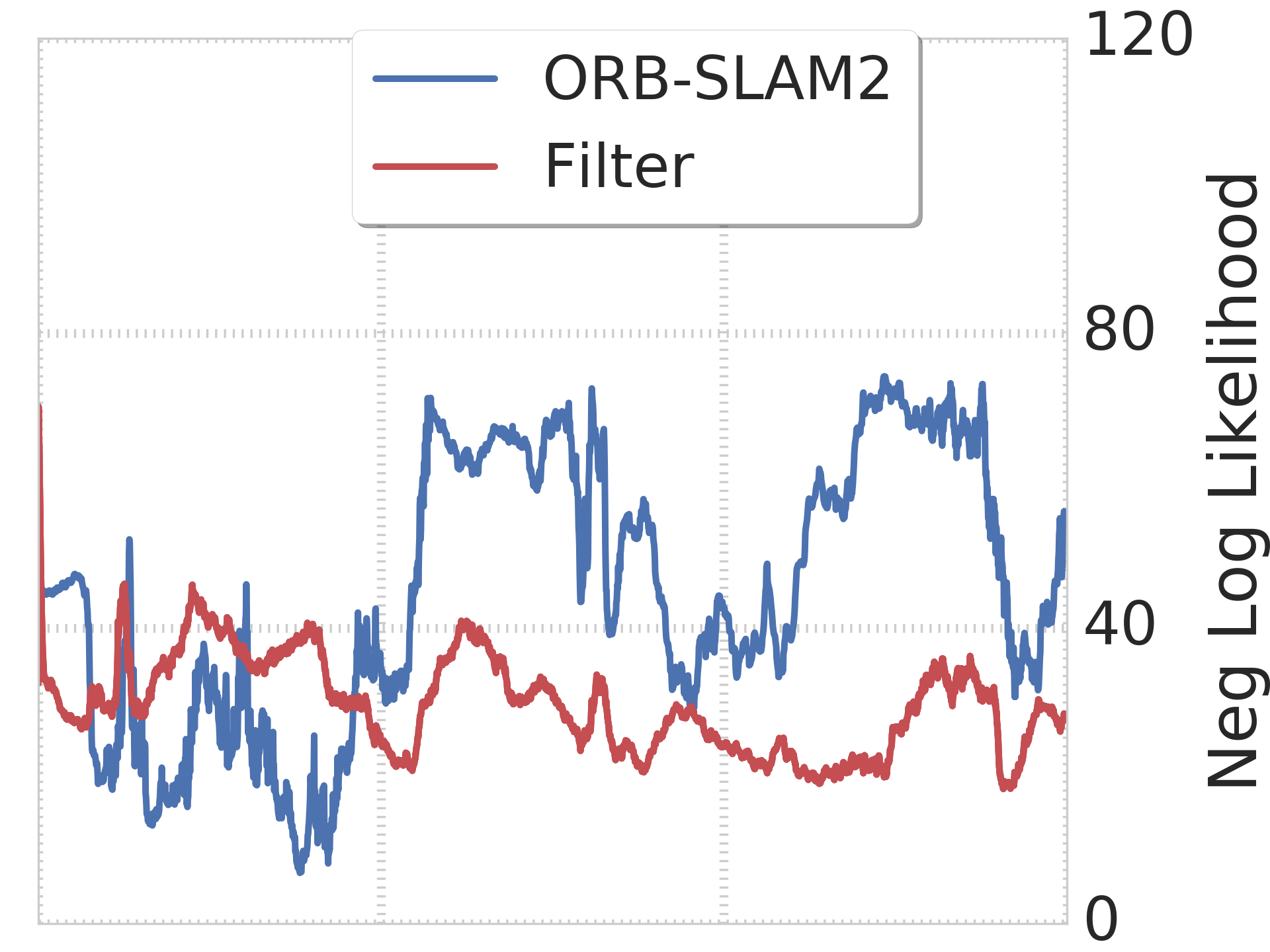} \\
		\includegraphics[width=0.45\linewidth]{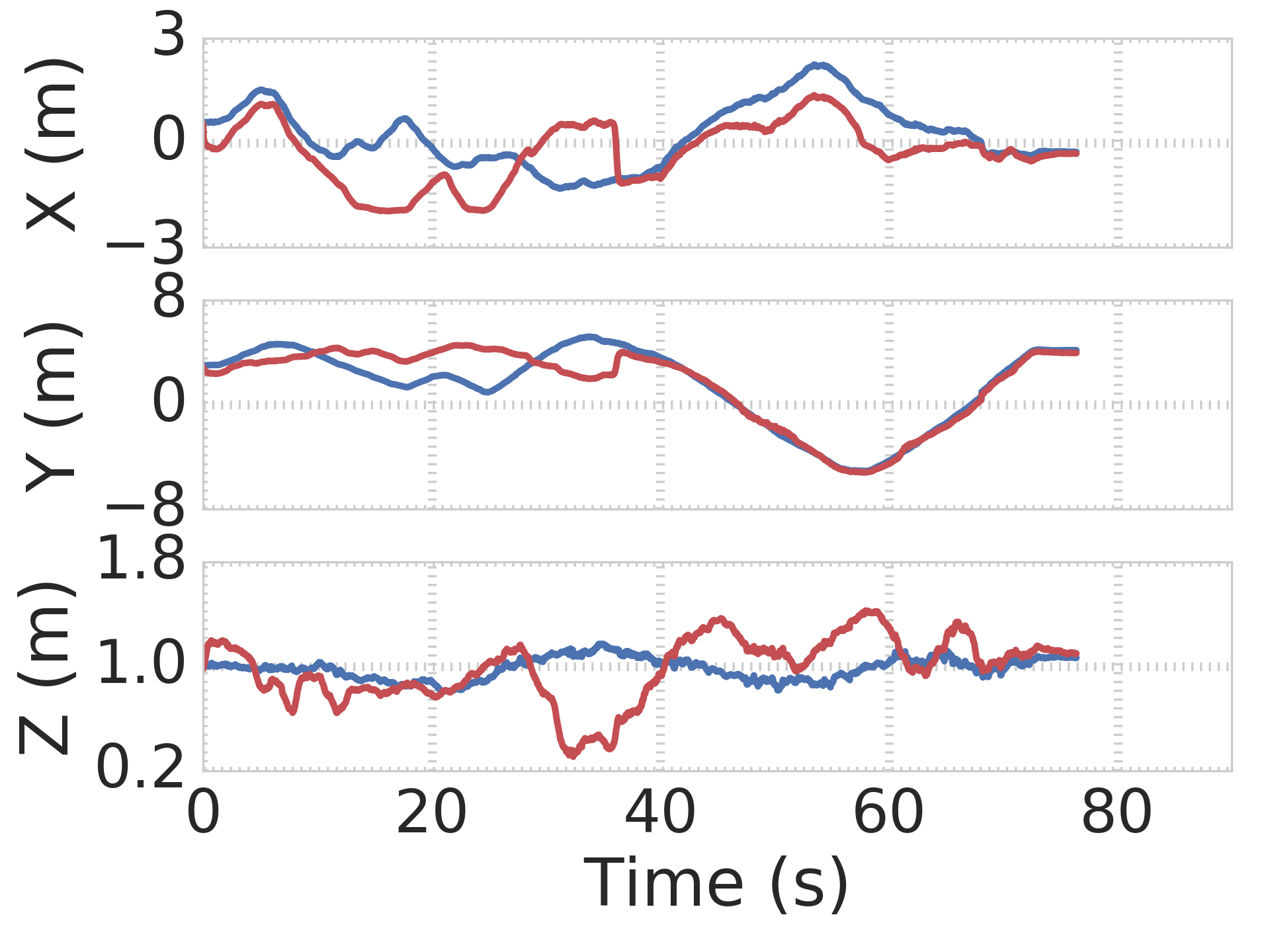}      &
		\includegraphics[width=0.45\linewidth]{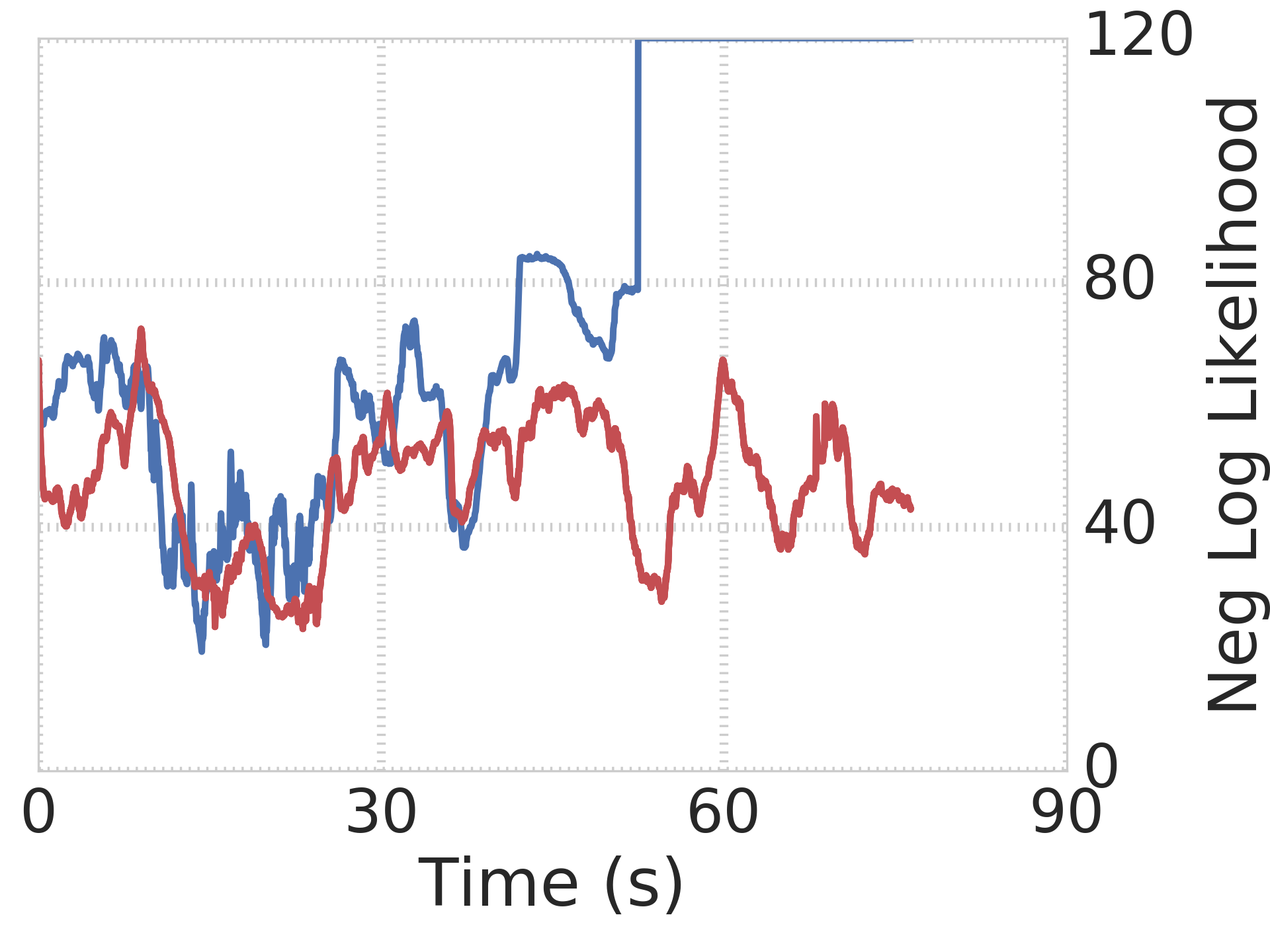}
	\end{tabular}
	\caption{Comparison between the position and corresponding likelihood estimates for two runs from ORB-SLAM2 and our filter, respectively. \textit{Top}: A nominal path with feature rich data, and \textit{Bottom}: A path moving through regions of low observability. Contrast the continually increasing divergence (capped in the graph) of the ORB-SLAM2 estimate after moving through the feature poor region with the lower snapped negative likelihood values for the same locations for our filter. The corresponding poses and overlaid depth scan at approximately 55s is shown in Fig.~\ref{fig:qualitative_comp}. Due to a minimal overlap of the depth scan with the map for the ORB-SLAM2 frame, the likelihood value is very low.}\label{fig:nsh_traj}
\end{figure}

\begin{figure}[h]
	\centering
	\includegraphics[width=0.6\linewidth]{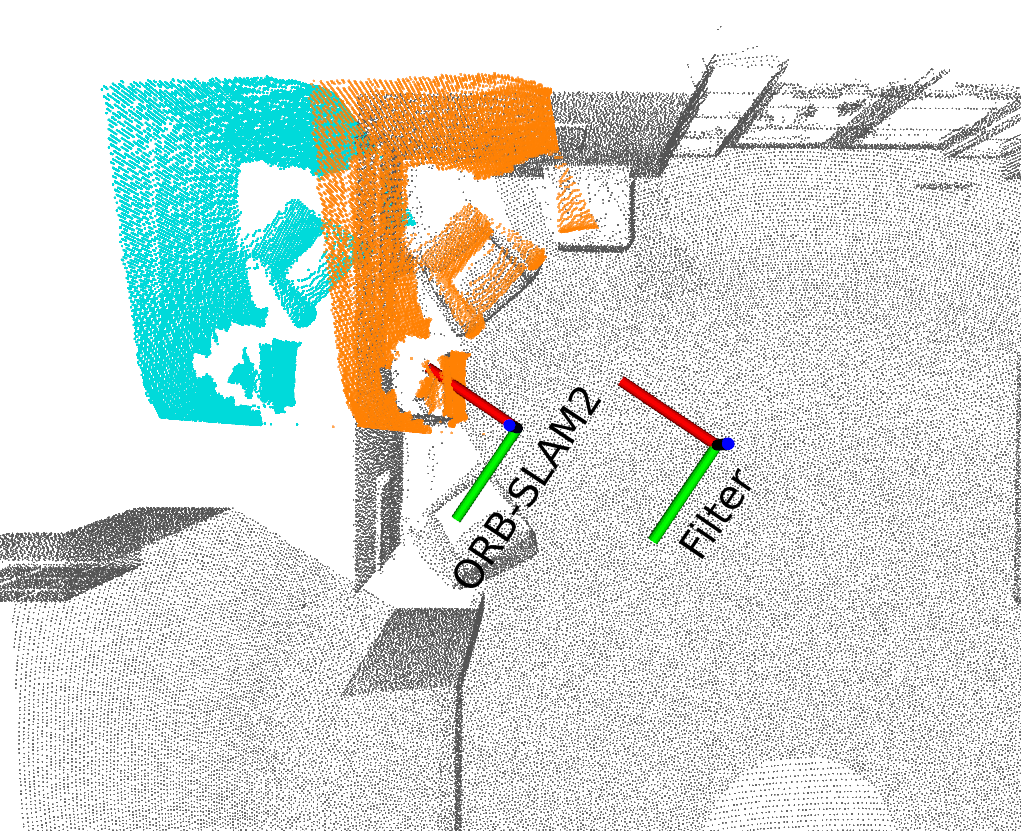}
	\caption{Comparison of registration of current sensor measurement at ground truth point cloud (gray) at ORB-SLAM2 pose estimation (cyan) and at the estimated filter pose (orange). The sensor measurement aligns with the ground truth point cloud in the filter estimate frame while the accumulated drift in the ORB-SLAM2 frame due to transition through a less feature rich region leads to poor alignment.}\label{fig:qualitative_comp}
\end{figure}

We generate a ground truth pointcloud using a FARO Focus 3D Laser scanner~\footnote{\url{https://www.faro.com/products/construction-bim-cim/faro-focus/}} and use an ASUS Xtion RGB-D camera for acquiring sensor data. An IMU strapped to the camera determines the roll and pitch of the sensor. For odometry, we only use the frame to frame relative transform as opposed to the global pose output from ORB-SLAM2 as input to the process model.
Note that the global ORB-SLAM2 position we compare to in Fig.~\ref{fig:nsh_traj} and Fig.~\ref{fig:qualitative_comp} is using loop closure to mitigate the drift in its frame to frame estimates.

As ground truth is not available for this dataset, we report the negative log likelihood values at the mean particle filter location and the reported ORB-SLAM2 poses. We show results of two runs in this environment in Fig.~\ref{fig:nsh_traj}: The first through a nominal path with feature rich data (as shown in detail earlier in Fig.~\ref{fig:mockup}) where the estimated positions of the sensor for the two approaches are very similar (but with worse likelihood values for ORB-SLAM2). The second run demonstrates the advantage of using particle filters over maximum likelihood estimators in that the former can converge to the correct result even after moving through a region of low observability. We observe that the sensor measurements register at the converged filter location better after snapping back than those for the ORB-SLAM2 estimate, as can be qualitatively seen in Fig.~\ref{fig:qualitative_comp}.

The particles in these experiments are initialized from a uniform distribution over a $4\text{m} \times 8\text{m} \times 3\text{m}$ for position and $\pi$ radians in yaw.

\subsubsection{Evaluation with TUM Dataset (\textbf{D4})}
To demonstrate the ability of our filter to generalize to both different odometry algorithms and datasets we compare the performance with three different odometry inputs as process models: The Generalized-ICP algorithm~\cite{segal2009generalized}, ORB-SLAM2 frame-to-frame relative transform, and ground truth odometry. The point cloud map of the environment was created by stitching several sensor scans together using their corresponding ground truth poses. In spite of the stitched point cloud not being as well registered as that from a FARO scanner due to sensor and ground truth pose noise, the performance of the filter is similar (Table~\ref{table:freiburg}).

\begin{table}[h]
	\centering
	\caption{Performance on D4 (RMSE in cm)}
	\scalebox{0.9}{
		\begin{tabular}{clccc}
			\toprule
			\multicolumn{3}{c}{Our Approach} & {ORB-SLAM2}                         \\
			Process Input                    & mean        & var ($cm^2$) &        \\
			\midrule
			{ORB-SLAM2 Velocity}             & 7.67        & 0.21         &        \\
			{Ground Truth Velocity}          & 7.56        & 0.28         & {4.55} \\
			{G-ICP Velocity}                 & 9.07        & 0.21         &        \\
			\bottomrule
		\end{tabular}}\label{table:freiburg}
\end{table}
\begin{figure}[h]
	\centering
	\includegraphics[width=0.8\linewidth]{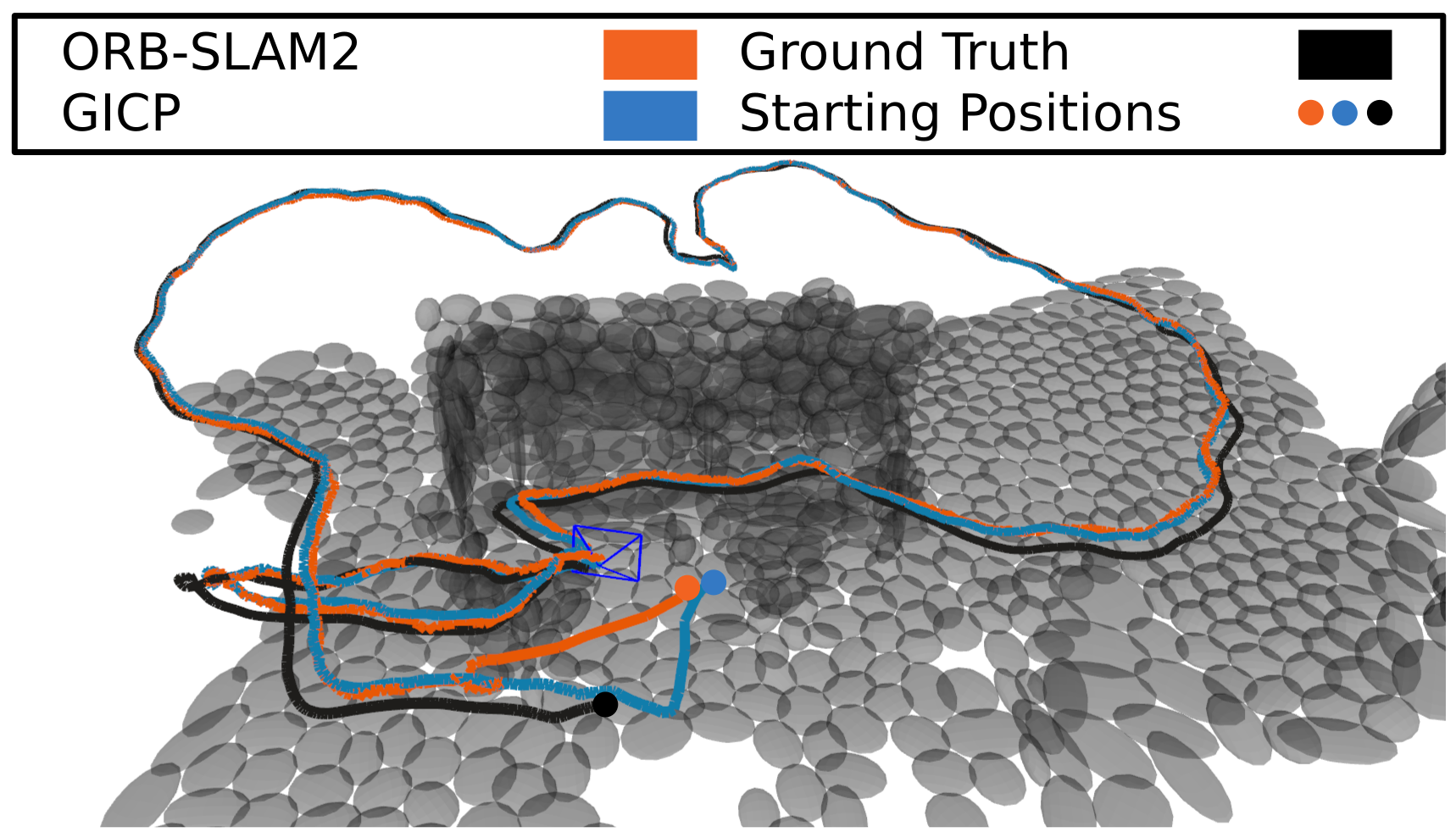}
	\caption{Comparison of our particle filter approach using ORB-SLAM2 frame-to-frame odometry (orange) and Generalized-ICP (cyan) as process models with ground truth pose (black) on TUM's Freiburg 3 Desk Dataset. The GMM representation of the world is created by stitching sensor scans using the ground truth pose estimates. The higher global error of our approach than that of ORB-SLAM2 can be attributed to the noisy reconstruction of the environment point cloud from the accumulated scans.}
\end{figure}

\subsection{Runtime Performance Analysis}
\begin{figure}[h]
	\centering
	\includegraphics[width=0.7\linewidth]{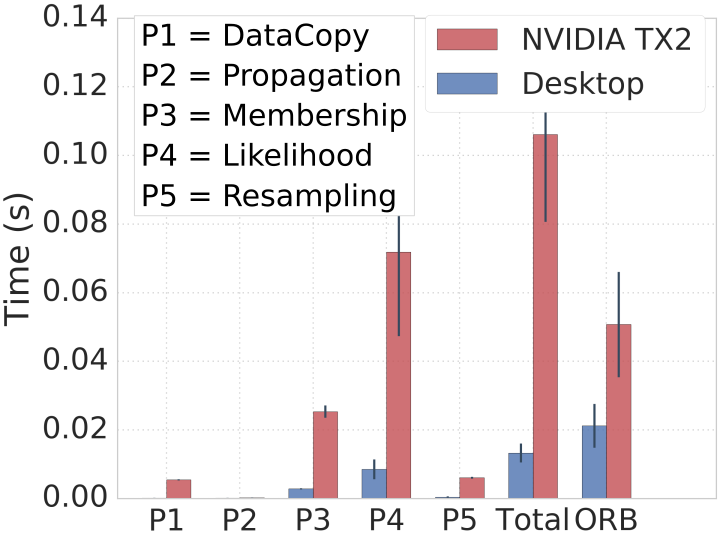}
	\caption{Execution time comparison for subcomponents of the algorithm for the D1(a) dataset on an Intel i7 desktop with an NVIDIA GPU and an embedded NVIDIA TX2 platform. Performance scales linearly with the number of CUDA cores. As a point of comparison ORB-SLAM2 runtime on the same dataset is faster on the embedded platform than on the desktop.}\label{fig:timing_plot}
\end{figure}

%
As seen in Fig.~\ref{fig:timing_plot}, the likelihood evaluation is the most computationally expensive operation. Execution time for this step varies with the number of Gaussian components used to compute likelihood for each image patch and therefore is dependent on the fidelity of the model.

\begin{figure}[h]
	\centering
	\includegraphics[width=0.7\linewidth]{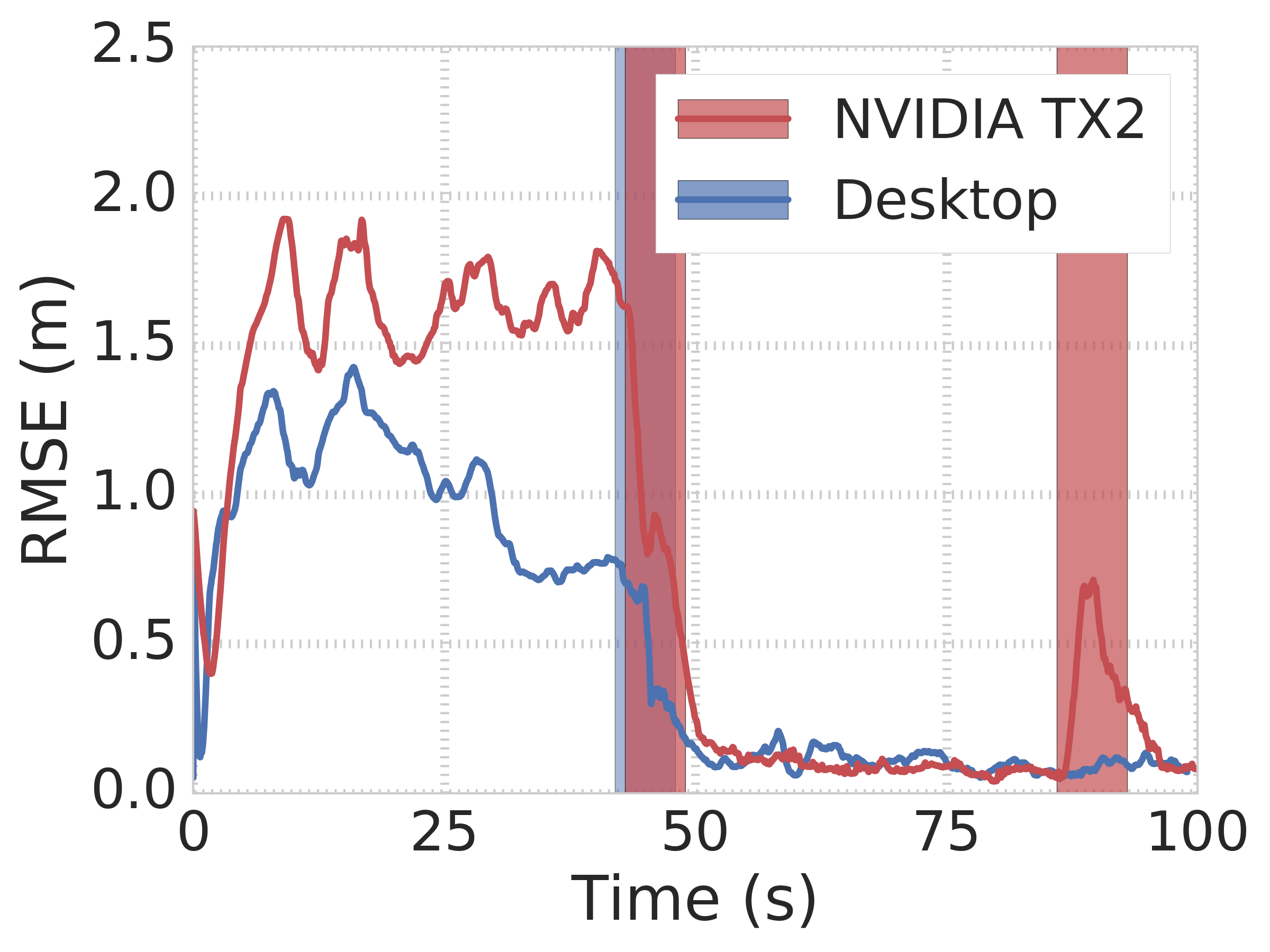}
	\caption{Comparison of the filter performance on the desktop with the NVIDIA TX2 on the D1(a) dataset. As the filter operates at a slower frame rate on the TX2 it initially exhibits a larger error but once the sensor observes a uniquely identifiable location, both trial sets converge to the ground truth location.}\label{fig:desktop_tx2_plot}
\end{figure}

The filter runs at an average rate of 80 Hz and 9.5 Hz on the Desktop and embedded class systems, respectively. This is comparable to the ORB-SLAM2 rates of 47 Hz and 20 Hz on the respective platforms. Initial convergence on the TX2 is slower due to the implicitly larger odometry steps. However, post convergence the metric performance is not significantly affected. As an illustrative example, the  impact of the slower runtime performance on the TX2 for D1(a) is demonstrated in Fig.~\ref{fig:desktop_tx2_plot}.

%% file: discussion.tex
We present a framework to perform real-time localization of depth sensors given a prior continuous spatial belief representation. Key to being able to do this is the ability to project a succinct representation into the image frame of the sensor to evaluate the likelihood of the data having originated from the given map for a given pose. By utilizing a fast likelihood computation approximation we can then perform robust particle filter localization in real-time even on an embedded GPU platform.

Despite the apparent suitability of the likelihood function to an optimization based approach for more fine grained pose refinement, it is not so straightforward. For Gaussian components that are more flat the gradient profile can vary rapidly in the vicinity of the components leading to poor conditioning that can challenge traditional iterative Gauss-Newton descent strategies. Further, the absence of strong gradient information in spatial data as encoded by the necessarily smooth-by-construction representation hinders the application of such methods.

Future steps will involve extending the formulation to hierarchical map representations to localize over larger scale environments. We also intend to investigate incorporation of this approach as a possible robust relocalization backend for a visual SLAM algorithm running onboard a relevant SWaP constrained aerial platform.

%% file: ms.bbl
\begin{thebibliography}{10}\itemsep=-1pt

\bibitem{angeli2008fast}
A.~Angeli, D.~Filliat, S.~Doncieux, and J.-A. Meyer.
\newblock {Fast and incremental method for loop-closure detection using bags of
  visual words}.
\newblock {\em IEEE Transactions on Robotics}, 24(5), 2008.

\bibitem{biswas2012depth}
J.~Biswas and M.~Veloso.
\newblock {Depth camera based indoor mobile robot localization and navigation}.
\newblock In {\em {Proc. of IEEE Intl. Conf. on Robotics and Automation}},
  2012.

\bibitem{bry2012state}
A.~Bry, A.~Bachrach, and N.~Roy.
\newblock {State estimation for aggressive flight in GPS-denied environments
  using onboard sensing}.
\newblock In {\em {Proc. of IEEE Intl. Conf. on Robotics and Automation}},
  2012.

\bibitem{burguera2008likelihood}
A.~Burguera, Y.~Gonz{\'a}lez, and G.~Oliver.
\newblock {The likelihood field approach to sonar scan matching}.
\newblock In {\em {Proc. of IEEE/RSJ Intl. Conf. on Intelligent Robots and
  Systems}}, 2008.

\bibitem{das20133d}
A.~Das, J.~Servos, and S.~L. Waslander.
\newblock {3D scan registration using the normal distributions transform with
  ground segmentation and point cloud clustering}.
\newblock In {\em {Proc. of IEEE Intl. Conf. on Robotics and Automation}},
  2013.

\bibitem{eckart2015mlmd}
B.~Eckart, K.~Kim, A.~Troccoli, A.~Kelly, and J.~Kautz.
\newblock {MLMD: Maximum Likelihood Mixture Decoupling for fast and accurate
  point cloud registration}.
\newblock In {\em {Proc. of IEEE Intl. Conf. on 3D Vision}}, 2015.

\bibitem{eckart2016accelerated}
B.~Eckart, K.~Kim, A.~Troccoli, A.~Kelly, and J.~Kautz.
\newblock {Accelerated generative models for 3D point cloud data}.
\newblock In {\em {Proc. of IEEE Conf. on Computer Vision and Pattern
  Recognition}}, 2016.

\bibitem{fallon2012efficient}
M.~F. Fallon, H.~Johannsson, and J.~J. Leonard.
\newblock {Efficient scene simulation for robust Monte Carlo localization using
  an RGB-D camera}.
\newblock In {\em {Proc. of IEEE Intl. Conf. on Robotics and Automation}},
  2012.

\bibitem{fang2015real}
Z.~Fang and S.~Scherer.
\newblock {Real-time onboard 6DoF localization of an indoor MAV in degraded
  visual environments using a RGB-D camera}.
\newblock In {\em {Proc. of IEEE Intl. Conf. on Robotics and Automation}},
  2015.

\bibitem{gonccalves2011real}
T.~Gon{\c{c}}alves and A.~I. Comport.
\newblock {Real-time Direct Tracking of Color Images in the Presence of
  Illumination Variation}.
\newblock In {\em {Proc. of IEEE Intl. Conf. on Robotics and Automation}},
  2011.

\bibitem{hershey2007approximating}
J.~R. Hershey and P.~A. Olsen.
\newblock {Approximating the Kullback Leibler divergence between Gaussian
  mixture models}.
\newblock In {\em {Proc. of the IEEE Intl. Conf. on Acoustics, Speech and
  Signal Processing}}, volume~4, 2007.

\bibitem{hornung2013octomap}
A.~Hornung, K.~M. Wurm, M.~Bennewitz, C.~Stachniss, and W.~Burgard.
\newblock {OctoMap: An efficient probabilistic 3D mapping framework based on
  octrees}.
\newblock {\em Autonomous Robots}, 34(3), 2013.

\bibitem{kitagawa1996monte}
G.~Kitagawa.
\newblock {{Monte Carlo filter and smoother for non-Gaussian nonlinear state
  space models}}.
\newblock {\em Journal of Computational and Graphical Statistics}, 5(1), 1996.

\bibitem{magnusson2015beyond}
M.~Magnusson, N.~Vaskevicius, T.~Stoyanov, K.~Pathak, and A.~Birk.
\newblock {Beyond points: Evaluating recent 3D scan-matching algorithms}.
\newblock In {\em {Proc. of IEEE Intl. Conf. on Robotics and Automation}},
  2015.

\bibitem{mur2017orb}
R.~Mur-Artal and J.~D. Tard{\'o}s.
\newblock {ORB-SLAM2: An Open-Source SLAM System for Monocular, Stereo, and
  RGB-D cameras}.
\newblock {\em IEEE Transactions on Robotics}, 33(5), 2017.

\bibitem{oishi2013nd}
S.~Oishi, Y.~Jeong, R.~Kurazume, Y.~Iwashita, and T.~Hasegawa.
\newblock {ND voxel localization using large-scale 3D environmental map and
  RGB-D camera}.
\newblock In {\em {Proc. of IEEE Intl. Conf. on Robotics and Biomimetics}},
  2013.

\bibitem{oksimultaneous}
K.~Ok, W.~N. Greene, and N.~Roy.
\newblock {{Simultaneous Tracking and Rendering: Real-time Monocular
  Localization for MAVs}}.
\newblock In {\em {Proc. of IEEE Intl. Conf. on Robotics and Automation}},
  2016.

\bibitem{oleynikova2016voxblox}
H.~Oleynikova, Z.~Taylor, M.~Fehr, R.~Siegwart, and J.~Nieto.
\newblock {Voxblox: Incremental 3D Euclidean Signed Distance Fields for
  On-Board MAV Planning}.
\newblock In {\em Proc. of IEEE/RSJ Intl. Conf. on Intelligent Robots and
  Systems}. IEEE, 2017.

\bibitem{pascoe2015robust}
G.~Pascoe, W.~Maddern, and P.~Newman.
\newblock {{Robust Direct Visual Localisation using Normalised Information
  Distance}}.
\newblock In {\em {Proc. of British Machine Vision Conf.}}, 2015.

\bibitem{segal2009generalized}
A.~Segal, D.~Haehnel, and S.~Thrun.
\newblock {Generalized-ICP}.
\newblock In {\em {Proc. of Robotics: Science and Systems}}, volume~2, 2009.

\bibitem{srivastava2016approximate}
S.~Srivastava and N.~Michael.
\newblock {Approximate Continuous Belief Distributions for Precise Autonomous
  Inspection}.
\newblock In {\em {Proc. of IEEE Intl. Symposium on Safety, Security, and
  Rescue Robotics}}, 2016.

\bibitem{stewart2012laps}
A.~D. Stewart and P.~Newman.
\newblock {{LAPS-Localisation using Appearance of Prior Structure: 6-DoF
  Monocular Camera Localisation using Prior Pointclouds}}.
\newblock In {\em {Proc. of IEEE Intl. Conf. on Robotics and Automation}},
  2012.

\bibitem{stoyanov2012fast}
T.~Stoyanov, M.~Magnusson, H.~Andreasson, and A.~J. Lilienthal.
\newblock {Fast and accurate scan registration through minimization of the
  distance between compact 3D NDT representations}.
\newblock {\em The Intl. Journal of Robotics Research}, 31(12), 2012.

\bibitem{sturm12iros}
J.~Sturm, N.~Engelhard, F.~Endres, W.~Burgard, and D.~Cremers.
\newblock In {\em {Proc. of IEEE/RSJ Intl. Conf. on Intelligent Robots and
  Systems},}.

\bibitem{thrun2005probabilistic}
S.~Thrun, W.~Burgard, and D.~Fox.
\newblock {\em {Probabilistic robotics}}.
\newblock MIT press, 2005.

\bibitem{van2001unscented}
R.~Van Der~Merwe, A.~Doucet, N.~De~Freitas, and E.~A. Wan.
\newblock {The unscented particle filter}.
\newblock In {\em {Advances in neural information processing systems}}, 2001.

\bibitem{zhou2013dense}
Q.-Y. Zhou and V.~Koltun.
\newblock {Dense scene reconstruction with points of interest}.
\newblock {\em ACM Transactions on Graphics (TOG)}, 32(4).

\end{thebibliography}
